\documentclass{article}
\usepackage{spconf,amsmath,graphicx}
\usepackage[colorlinks=true, allcolors=blue]{hyperref}
\usepackage{array}
\usepackage{booktabs}
\usepackage{caption}
\usepackage{subcaption}
\usepackage{pgf}
\usepackage{import}
\usepackage{layouts}

\title{Neural information coding for efficient spike-based image denoising}
\name{Andrea Castagnetti, Alain Pegatoquet, Benoît Miramond\thanks{This work has been supported by the French governement through 3IA Côte d'Azur institute, reference ANR-19-P3IA-0002}}
\address{Université Côte d'Azur, CNRS, LEAT\\
        firstname.lastname@univ-cotedazur.fr}
\begin{document}
%
\maketitle
\begin{abstract}
In recent years, Deep Convolutional Neural Networks (DCNNs) have outreached the performance of classical algorithms for image restoration tasks. However most of these methods are not suited for computational efficiency and are therefore too expensive to be executed on embedded and mobile devices. In this work we investigate Spiking Neural Networks (SNNs) for Gaussian denoising, with the goal of approaching the performance of conventional DCNN while reducing the computational load. We propose a formal analysis of the information conversion processing carried out by the Leaky Integrate and Fire (LIF) neurons and we compare its performance with the classical rate-coding mechanism. The neural coding schemes are then evaluated through experiments in terms of denoising performance and computation efficiency for a state-of-the-art deep convolutional neural network. Our results show that SNNs with LIF neurons can provide competitive denoising performance but at a reduced computational cost.
\end{abstract}

\section{Introduction and Related Work}
\label{sec:intro}
Smartphone cameras, because of their reduced size and high pixel count, are intrinsically more susceptible to noise than conventional digital cameras. Image denoising algorithms are therefore intensively used in smartphones to recover image quality by reducing the amount of noise of the raw image.
Image denoising performance have increased during the last few years and recent methods based on Deep Convolutional Neural Networks (DCNNs) have provided very high scores \cite{zhang_beyond_2017} to the point of outreaching classical spatial and patch-based algorithms \cite{gu_weighted_2014}.
However, deploying AI-based algorithms on embedded devices poses many problems. The limited amount of memory available, power consumption and thermal dissipation are indeed critical for embedded battery powered platforms.
The field of neuromorphic engineering, especially SNNs, is emerging as a new paradigm for the design of low-power and real-time information processing hardware \cite{ABDERRAHMANE2020366}. The spike information coding used by SNNs enables sparse and event-based computation through the network. The combination of these properties may lead to more energy efficient hardware implementations of neural networks, allowing state-of-the-art AI algorithms to be executed on mobile platforms with a reduced power budget \cite{mendez2022edge}. However, to achieve these energy gains while simultaneously reaching the level of performance of DCNNs, SNNs must be able to encode analog data with high precision using very compact codes, i.e. spike trains. In recent years, several training and conversion methods have been proposed to improve the accuracy of SNNs on large-scale machine learning tasks. 
To take advantage of better performance provided by supervised learning, several methods have been developed to convert ANNs, trained using standard schemes like backpropagation, into SNNs for event-driven inference \cite{diehl_fast-classifying_2015} \cite{rueckauer_conversion_2017}. The ANN-SNN conversion is based on the idea that firing rates of spiking neurons should match the activations of analog neurons. Rate-based conversion methods have achieved significant results over the last few years, thus reducing the accuracy gap with ANNs \cite{sengupta_going_2019}. However, rate-based conversion methods have a major drawback since they require a large amount of timesteps to precisely match the activations of analog neurons. A new approach, called surrogate gradient learning, has been proposed to train SNNs directly in the spike domain using standard supervised learning algorithms \cite{neftci_surrogate_2019}. Recent studies reported competitive performance on a series of static and dynamic datasets using surrogate gradient training \cite{wozniak_deep_2020}. 
In this paper, we will extend these previous works and study the trade off between accuracy and efficiency of SNNs for the specific and uncovered case of image denoising. This task is challenging for two reasons. First, as denoising is a regression task, the network has to predict a continuous value (i.e. the noise amplitude) for each pixel of the image. Moreover, state of the art results have been obtained with very deep networks (17 layers or more). In Section \ref{sec:neural_coding} we study two spike coding approaches and we formalize the trade-off between performance and activation sparsity in SNN. In Section \ref{sec:image_denoising} we propose, for the first time, an image denoising solution based on SNN. The network trained directly in the spike domain provides a level of performance close to the state of the art CNN based solution. In the last section, we conclude the paper and we discuss future work.

\section{Neural coding}
\label{sec:neural_coding}
Neural coding schemes convert input pixels into spikes that are transmitted to spiking neurons for information processing.
Specifically, we are interested in the two complementary neural coding procedures called \emph{encoding} and \emph{decoding} that map analog values into train of spikes and its inverse. Two types of encoding and decoding schemes are studied: the coding with information conversion and the rate-coding.
The following section presents an analysis of the neural conversion with LIF neurons. The comparison between the two coding scheme will be discussed in Sec. \ref{subsec:image_coding_lif}

\subsection{Neural coding with LIF neurons}
\label{subsec:neural_coding_lif}
In this first coding scheme, that we call \emph{Neural information conversion}, a LIF neuron, located in the first layer (encoding) of the network is fed with a constant input $x$ through $T$ timesteps. We are interested in finding the encoding function that defines the relation between $x$ and $z(t)$, the spiking output. Let us first recall the equations that govern the dynamic of a LIF neuron in the discrete case \cite{doutsi_dynamic_2021}:
\begin{equation}
V[n] = V[n-1] + \frac{1}{\tau} (-(V[n-1] - V_{reset}) + x[n])
\label{eq:lif_V_discrete}
\end{equation}
Whenever $V[n] \geq V_{th}$, where $V_{th}$ is the threshold voltage, the neuron emits a binary spike. The spike train representation of the analog input $x$, is thus encoded in the following function:
\begin{equation}
z(t) = \sum_{j=1}^{T} \delta(t - t_j)
\label{eq:z_train_spikes}
\end{equation}
Where $\delta(t)$ is the Dirac delta function and $t_j$ are the spike-times indexed by $j$.
In the following we also consider that the membrane potential is completely discharged after a spike emission, $V_{reset} = 0$. Let us find the value of $x$ that makes the neuron fires at each timestep, thus producing a constant firing rate of 1. The conditions that produce this firing pattern are shown below:
\begin{equation}
    \begin{cases}
      V[n - 1] = 0\\
      V[n] \geq V_{th}\\
      x[n] = x, \text{$\forall \, n$}
    \end{cases} 
\label{eq:lif_cond_fr_1}
\end{equation}
Since a spike has to be generated at the current timestep $n$, the membrane potential is greater or equal to $V_{th}$. Moreover, with a firing rate of 1, the neuron resets its membrane potential at each timestep, after the spike emission. Therefore the membrane potential at the previous timestep, $V[n - 1]$, equals 0. 
Replacing conditions \ref{eq:lif_cond_fr_1} into Eq. \ref{eq:lif_V_discrete} we obtain:
\begin{equation}
x \geq V_{th} . \tau
\label{eq:lif_x_fr_1}
\end{equation}
When the input $x$ is greater than or equal to $V_{th} . \tau$, the LIF neuron fires a spike at each timestep. Following the same reasoning, let us find the values of $x$ that produce a firing rate of 0.5.
In such a case, the neuron periodically alternates between two states: [charge, fire\&reset, charge, fire\&reset, \ldots].
The conditions that lead to this behaviour are shown below:
\begin{equation}
    \begin{cases}
      V[n-1] = x/\tau\\
      V[n] \geq V_{th}\\
      x[n] = x, \text{$\forall \, n$}
    \end{cases} 
\label{eq:lif_cond_fr_50}
\end{equation}
Replacing conditions \ref{eq:lif_cond_fr_50} into Eq. \ref{eq:lif_V_discrete} we obtain:
\begin{equation}
V[n] = V_{th} = x/\tau + \frac{1}{\tau} (x - x/\tau)
\label{eq:lif_x_fr_50_a}
\end{equation}
The values of $x$ that produces a firing rate of 0.5 are defined by:
\begin{equation}
 \frac{V_{th}}{ 2/\tau - 1/\tau^2} \leq x \leq V_{th}\, . \tau
\label{eq:lif_x_fr_50_b}
\end{equation}

\begin{figure}[htb]
\includegraphics[width=8.5cm]{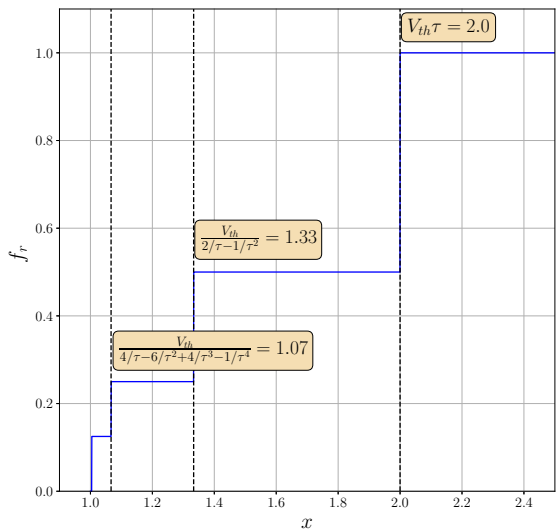}
\caption{Firing-rate as a function of the input $x$ for $T=8$ ($V_{th}=1.0, \tau=2.0$). The thresholds for $f_r = 1.0$ and $f_r = 0.5$ and $f_r = 0.25$ are also shown.}
\label{fig:fr_lif_encoding}
\end{figure}

The same approach can be used to determine the production conditions of the other firing rates of a LIF model as depicted in figure \ref{fig:fr_lif_encoding}.\\
Before proceeding in the analysis, it is now interesting to focus on the spike patterns generated by a LIF neuron in more details. We may wonder, for example, if all spiking patterns of a LIF neuron are allowed, as well as the number of different firing rates.
Let us start with a simple example where the neuron codes information over $T=8$ timesteps. Note that a sequence that leads to a generation of a spike (in the case of a constant input) must have the following format: [charge] during $k$ timesteps, followed by [fire\&reset].
Table \ref{tab:lif_fr_table_T8}, shows the spiking pattern generated by a LIF neuron, when simulated for $T=8$ timesteps.
\begin{table}
\centering
\begin{tabular}{cccccccc|c|c}
\toprule
\multicolumn{8}{c|}{Timesteps} & $f_r$ & $k$\\
0 & 1 & 2 & 3 & 4 & 5 & 6 & 7 & {} & {}\\
\midrule
1 & 1 & 1 & 1 & 1 & 1 & 1 & 1 & {1.0} & 0\\
0 & 1 & 0 & 1 & 0 & 1 & 0 & 1 & {0.5} & 1\\
0 & 0 & 1 & 0 & 0 & 1 & 0 & 0 & {0.25} & 2\\
0 & 0 & 0 & 1 & 0 & 0 & 0 & 1 & {0.25} & 3\\
0 & 0 & 0 & 0 & 1 & 0 & 0 & 0 & {0.125} & 4\\
0 & 0 & 0 & 0 & 0 & 1 & 0 & 0 & {0.125} & 5\\
0 & 0 & 0 & 0 & 0 & 0 & 1 & 0 & {0.125} & 6\\
0 & 0 & 0 & 0 & 0 & 0 & 0 & 1 & {0.125} & 7\\
0 & 0 & 0 & 0 & 0 & 0 & 0 & 0 & {0} & 8\\
\bottomrule
\end{tabular}
\caption{\label{tab:lif_fr_table_T8} Output codes of a LIF neuron stimulated with a constant value. $T$ represents the number of timesteps (here $T=8$). $f_r$ is the firing rate and $k$ is the number of timesteps used for the charge phase before the first spike. The value 1 means that a spike has been generated on the corresponding timestep.}
\end{table}
We can observe that the number of output codes does not depend on the input $x$ but only on the value of $T$ (simulation timesteps). In fact, only $T+1$ output codes can be generated by a LIF neuron, when stimulated with a constant input. We have so far characterized the \emph{encoding} function, the relation between the analog input $x$ and the spike pattern, $z(t)$, generated at the output of the neuron. The inverse process called \emph{decoding} aims at mapping the inverse function, that is the reconstruction of the analog value ($\hat{x}$) from a spiking input. To do so, we use the firing-rate of the neuron as the information carrying variable and express the decoded analog output as follows, where $f_r$ denotes the firing rate:
\begin{equation}
\hat{x} = f_r =  \frac{1}{T}\sum^{T} z(t)
\label{eq:lif_x_decoded}
\end{equation}

The output codes obtained by simulating a LIF neuron are plotted in Fig. \ref{fig:fr_lif_encoding} as a function of the input value $x$.
As it can be observed, the conversion operated by a LIF neuron is highly non-uniform as it provides more quantization steps at amplitudes near $V_{th}$ than at higher amplitudes. However, the quantization step sizes decrease while approaching $V_{th}$. At the opposite, codes that carry a high $f_r$, have large quantization step sizes. As an example and as shown in Fig. \ref{fig:fr_lif_encoding}, a $f_r$ of 0.5 will be generated by the neuron when $x \in [1.33, 2.0]$ .

\subsection{Comparison between neural conversion and rate coding}
\label{subsec:image_coding_lif}
\begin{figure}[htb]
\includegraphics[width=8.5cm]{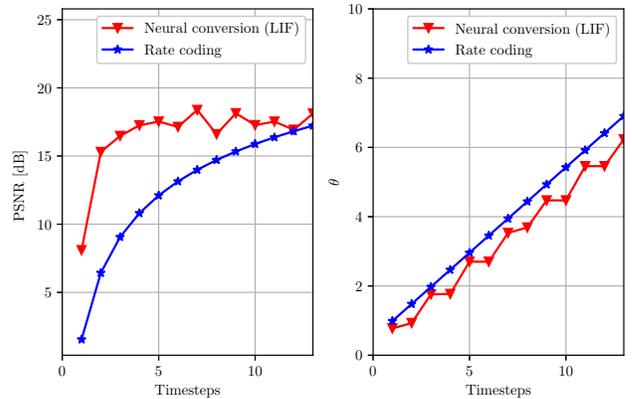}
\caption{PSNR (left) and $\theta$ (right) as a function of the number of timesteps for the LIF neural conversion and rate coding schemes on the Set12 dataset. Each point of the curves represents an average over the number of pixels of all the dataset images. Here $V_{th} = 1$ and $\tau = 2$ for the LIF neurons.}
\label{fig:psnr_theta_set12}
\end{figure}

To assess the performance of the neural conversion scheme, we use a set of 12 natural images \cite{roth_fields_2005} (Set12), to measure the accuracy of the quantizer. The Peak-Signal-to-Noise ratio (PSNR) is used as quality criterion.
The pixel intensities of the test images are normalized in the interval $[1.0, 2.0]$ to match the conversion range of the LIF neuron shown in Fig. \ref{fig:fr_lif_encoding}. The normalized pixel intensities are fed (without noise) into a LIF neuron membrane for $T$ timesteps. The spikes generated by each neuron are then collected and an estimate of each pixel value is computed using equation \ref{eq:lif_x_decoded}. We compare the neural coding of LIF neurons with the rate coding, a well known scheme that has been extensively used in the SNN community for coding dense information in the spike domain \cite{guo_neural_2021} \cite{abderrahmane_neural_2020} \cite{brette_philosophy_2015}. In the rate coding scheme we assume that spike trains are independent realizations of Poisson processes with rates $r_i$, where the pixel intensity $x_i$ is the firing probability, normalized between [0,1], at each timestep. Eq. \ref{eq:lif_x_decoded} is also used to decode the spiking output. 
The average PSNR on Set12 is shown on the left side of Fig. \ref{fig:psnr_theta_set12}.
As we can observe, the PSNR increases quickly for the LIF coding scheme and saturates at $T\sim10$. Adding more timesteps to the conversion does not improve the image reconstruction. From the previous analysis, presented in Sec. \ref{subsec:neural_coding_lif}, we know that increasing $T$ adds more quantization intervals and therefore also increases the number of output codes that can be generated by a LIF neuron. However, as it can be observed from Fig. \ref{fig:fr_lif_encoding}, the sizes of newly added quantized intervals decrease fast and vanish when $x$ approaches $V_{th}$. As a result, the non-uniform quantization scheme that emerges from the LIF neuron does not allow decreasing the distortion between the original and the quantized signals by adding more quantization bits, i.e. increasing $T$. On the other hand, the rate conversion scheme does not set any limits on the accuracy since the PSNR increases with a logarithmic shape as a function of $T$. Let us now study a property of great interest for a neural information coding scheme: the activity of the spiking neurons. This property is key for reducing computation costs, thus the energy consumption in networks of spiking neurons. 
The activity of a neural network is defined as the average number of spikes generated by each neuron during $T$ timesteps. Referring to the previous image quantization example, the activity is defined as follows:
\begin{equation}
\theta = \frac{\displaystyle\sum_{i=0}^{n} \sum_{j=0}^{m} z_{i,j}(t)}{n\times m}
\label{eq:lif_sparsity}
\end{equation}
Where $(n,m)$ is the size of the input image.
Summing over $z$, which is a $T \times n \times m$ binary matrix, results in the total number of spikes generated by all the LIF neurons. The activity, $\theta$, of the
rate coding and LIF conversion scheme can be seen on the right side of Fig. \ref{fig:psnr_theta_set12}.
This figure shows that, like in each rate conversion scheme, the number of spikes increases almost linearly with $T$. However, as we can observe from the PSNR curve shown in Fig. \ref{fig:psnr_theta_set12}, the amount of information carried by each new spike in the neural conversion scheme saturates above $T \sim 10$. The process of coding dense information into stream of spikes is key for SNN and has been pointed out as one of the main reasons for the current performance gap between SNN and ANN. 
In the next section we investigate how these rules and properties show up in larger and complex networks of spiking neurons.


\section{Image denoising with spiking neurons}
\begin{figure}[htb]
\includegraphics[width=8.5cm]{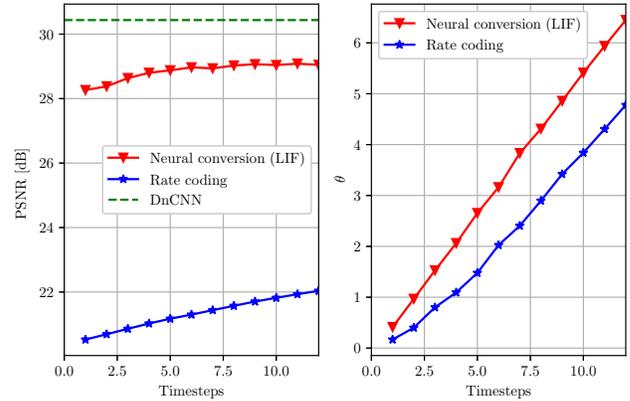}
\caption{Denoising PSNR (left) and $\theta$ (right) as a function of the number of conversion timesteps for the LIF neural conversion scheme and the rate coding. The dotted line represent the performance of DnCNN for a noise level $\sigma=25$.}
\label{fig:psnr_theta_lif_denoising}
\end{figure}

\begin{figure}[!htb]
\begin{minipage}[b]{0.48\linewidth}
  \centering
  \centerline{\includegraphics[width=2.4cm]{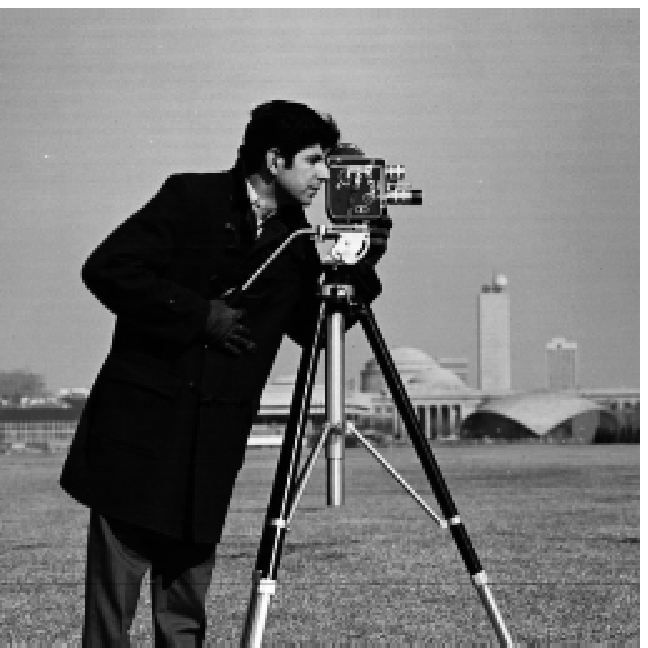}}
  \centerline{(a) Clean image}\medskip
\end{minipage}
\hfill
\begin{minipage}[b]{.48\linewidth}
  \centering
  \centerline{\includegraphics[width=2.4cm]{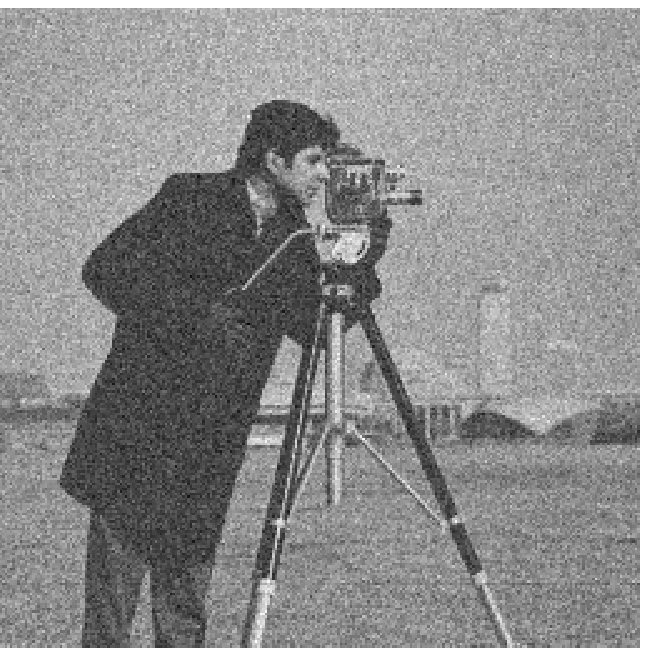}}
  \centerline{(b) Noisy image (20.18 dB)}\medskip
\end{minipage}
\begin{minipage}[b]{0.48\linewidth}
  \centering
  \centerline{\includegraphics[width=2.4cm]{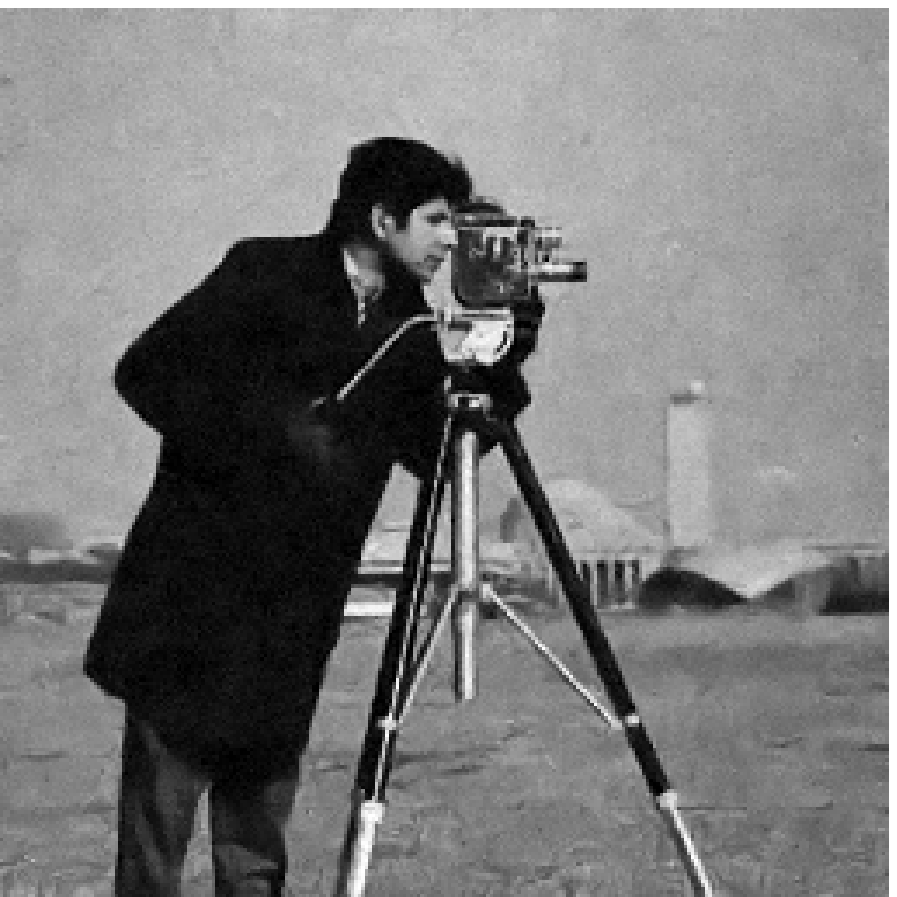}}
  \centerline{(c) LIF (28.35 dB)}\medskip
\end{minipage}
\hfill
\begin{minipage}[b]{.48\linewidth}
  \centering
  \centerline{\includegraphics[width=2.4cm]{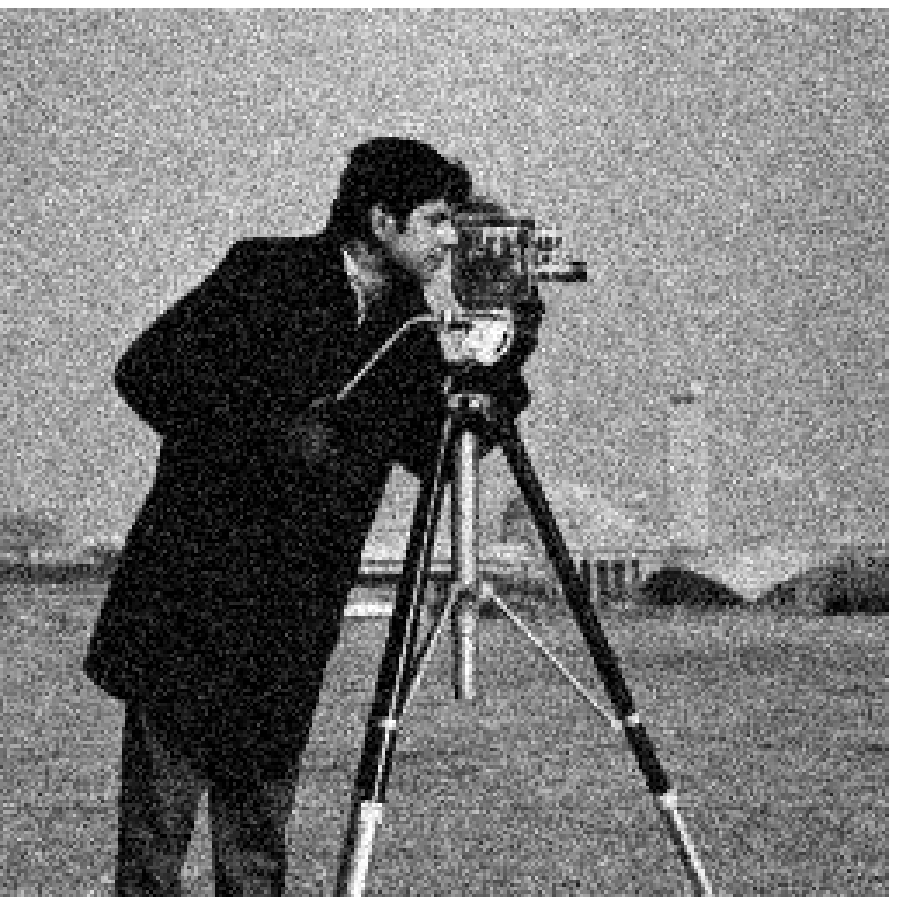}}
  \centerline{(d) Rate coding (21.59 dB)}\medskip
\end{minipage}

\caption{Denoising results of the image “C.man” with noise level 25. Denoised images are shown in Fig. c and d for LIF conversion and rate coding, both with $T=7$. 
}
\label{fig:cman_results}
\end{figure}

Our study of image denoising with spiking neurons is based on the DnCNN network proposed in \cite{zhang_beyond_2017}. We focus the study on Gaussian denoising with a certain noise level. The considered network is composed of 17 convolutional layers. Activation functions are replaced with LIF neurons ($V_{th} = 1$, $\tau = 2$). The input of the network is a noisy observation $y = x + v$ where $v$ is additive white Gaussian noise with zero mean and standard deviation $\sigma$. As proposed in \cite{zhang_beyond_2017} we follow a residual learning formulation, to train a mapping $R(y) \sim v$ and recover the clean image from $x = y - R(y)$. The averaged mean square error between the desired and the estimated residuals is used as the loss function. Training data for gaussian gray image denoising are generated using the same method proposed in \cite{zhang_beyond_2017}. 
A dataset composed of 12 images (not included in the train set) is used for testing. The surrogate gradient \cite{neftci_surrogate_2019} learning is used to train the SNN networks. Denoising results and neuron activity are shown in Fig. \ref{fig:psnr_theta_lif_denoising}. As we can observe the network performance (PSNR) follows the same trend observed for information coding shown in Fig. \ref{fig:psnr_theta_set12}. The LIF conversion scheme can provide competitive denoising performance with few timesteps, but PSNR saturates when $T>10$. Rate coding could theoretically achieve the same PSNR as DnCNN but at the cost of hundreds of timesteps.
\label{sec:image_denoising}
Fig. \ref{fig:cman_results} illustrates the visual results of the coding methods on the C.man image. As it can be seen with only 7 timesteps, the LIF conversion method provides better images compared with rate coding. The latter scheme would require a large amount of timesteps to encode analog values with the precision needed for the denoising task.

\section{Conclusion}
\label{sec:conclusion}
In this paper we have presented an analysis, based on information coding, for SNNs and its application for image denoising. Our analysis showed that information coding at the neuron level can explain the performance at the network level. As future work, we aim at using our approach to guide the design of spiking neural models. Our objective is to encode information with both low latency and high precision for further hardware neuromorphic implementation.

\vfill\pagebreak



\bibliographystyle{IEEEbib}
\bibliography{main.bib}

\begin{thebibliography}{10}

\bibitem{zhang_beyond_2017}
Kai Zhang, Wangmeng Zuo, Yunjin Chen, Deyu Meng, and Lei Zhang,
\newblock ``Beyond a {Gaussian} {Denoiser}: {Residual} {Learning} of {Deep}
  {CNN} for {Image} {Denoising},''
\newblock {\em IEEE Transactions on Image Processing}, vol. 26, no. 7, pp.
  3142--3155, July 2017,
\newblock arXiv: 1608.03981.

\bibitem{gu_weighted_2014}
Shuhang Gu, Lei Zhang, Wangmeng Zuo, and Xiangchu Feng,
\newblock ``Weighted {Nuclear} {Norm} {Minimization} with {Application} to
  {Image} {Denoising},''
\newblock in {\em 2014 {IEEE} {Conference} on {Computer} {Vision} and {Pattern}
  {Recognition}}, Columbus, OH, USA, June 2014, pp. 2862--2869, IEEE.

\bibitem{ABDERRAHMANE2020366}
Nassim Abderrahmane, Edgar Lemaire, and Benoît Miramond,
\newblock ``Design space exploration of hardware spiking neurons for embedded
  artificial intelligence,''
\newblock {\em Neural Networks}, vol. 121, pp. 366--386, 2020.

\bibitem{mendez2022edge}
Javier Mendez, Kay Bierzynski, MP~Cu{\'e}llar, and Diego~P Morales,
\newblock ``Edge intelligence: Concepts, architectures, applications and future
  directions,''
\newblock {\em ACM Transactions on Embedded Computing Systems (TECS)}, 2022.

\bibitem{diehl_fast-classifying_2015}
Peter~U. Diehl, Daniel Neil, Jonathan Binas, Matthew Cook, Shih-Chii Liu, and
  Michael Pfeiffer,
\newblock ``Fast-classifying, high-accuracy spiking deep networks through
  weight and threshold balancing,''
\newblock in {\em 2015 {International} {Joint} {Conference} on {Neural}
  {Networks} ({IJCNN})}, July 2015, pp. 1--8,
\newblock ISSN: 2161-4407.

\bibitem{rueckauer_conversion_2017}
Bodo Rueckauer, Iulia-Alexandra Lungu, Yuhuang Hu, Michael Pfeiffer, and
  Shih-Chii Liu,
\newblock ``Conversion of {Continuous}-{Valued} {Deep} {Networks} to
  {Efficient} {Event}-{Driven} {Networks} for {Image} {Classification},''
\newblock {\em Frontiers in Neuroscience}, vol. 11, pp. 682, 2017.

\bibitem{sengupta_going_2019}
Abhronil Sengupta, Yuting Ye, Robert Wang, Chiao Liu, and Kaushik Roy,
\newblock ``Going {Deeper} in {Spiking} {Neural} {Networks}: {VGG} and
  {Residual} {Architectures},''
\newblock {\em Frontiers in Neuroscience}, vol. 13, pp. 95, 2019.

\bibitem{neftci_surrogate_2019}
Emre~O. Neftci, Hesham Mostafa, and Friedemann Zenke,
\newblock ``Surrogate {Gradient} {Learning} in {Spiking} {Neural} {Networks}:
  {Bringing} the {Power} of {Gradient}-{Based} {Optimization} to {Spiking}
  {Neural} {Networks},''
\newblock {\em IEEE Signal Processing Magazine}, vol. 36, no. 6, pp. 51--63,
  Nov. 2019.

\bibitem{wozniak_deep_2020}
Stanisław Woźniak, Angeliki Pantazi, Thomas Bohnstingl, and Evangelos
  Eleftheriou,
\newblock ``Deep learning incorporating biologically inspired neural dynamics
  and in-memory computing,''
\newblock {\em Nature Machine Intelligence}, vol. 2, no. 6, pp. 325--336, June
  2020.

\bibitem{doutsi_dynamic_2021}
Effrosyni Doutsi, Lionel Fillatre, Marc Antonini, and Panagiotis Tsakalides,
\newblock ``Dynamic {Image} {Quantization} {Using} {Leaky}
  {Integrate}-and-{Fire} {Neurons},''
\newblock {\em IEEE Transactions on Image Processing}, vol. 30, pp. 4305--4315,
  2021.

\bibitem{roth_fields_2005}
S.~Roth and M.J. Black,
\newblock ``Fields of {Experts}: a framework for learning image priors,''
\newblock in {\em 2005 {IEEE} {Computer} {Society} {Conference} on {Computer}
  {Vision} and {Pattern} {Recognition} ({CVPR}'05)}, June 2005, vol.~2, pp.
  860--867 vol. 2,
\newblock ISSN: 1063-6919.

\bibitem{guo_neural_2021}
Wenzhe Guo, Mohammed~E. Fouda, Ahmed~M. Eltawil, and Khaled~Nabil Salama,
\newblock ``Neural {Coding} in {Spiking} {Neural} {Networks}: {A} {Comparative}
  {Study} for {Robust} {Neuromorphic} {Systems},''
\newblock {\em Frontiers in Neuroscience}, vol. 15, pp. 638474, Mar. 2021.

\bibitem{abderrahmane_neural_2020}
Nassim Abderrahmane and Benoit Miramond,
\newblock ``Neural coding: adapting spike generation for embedded hardware
  classification,''
\newblock in {\em 2020 {International} {Joint} {Conference} on {Neural}
  {Networks} ({IJCNN})}, Glasgow, United Kingdom, July 2020, pp. 1--8, IEEE.

\bibitem{brette_philosophy_2015}
Romain Brette,
\newblock ``Philosophy of the {Spike}: {Rate}-{Based} vs. {Spike}-{Based}
  {Theories} of the {Brain},''
\newblock {\em Frontiers in Systems Neuroscience}, vol. 9, pp. 151, 2015.

\end{thebibliography}

\end{document}